\documentclass{article}
\usepackage{graphicx}
\graphicspath{{figs/}}

\usepackage{arxiv}

\usepackage[utf8]{inputenc} 
\usepackage[T1]{fontenc}    
\usepackage{hyperref}       
\usepackage{url}            
\usepackage{booktabs}       
\usepackage{amsfonts}       
\usepackage{nicefrac}       
\usepackage{microtype}      
\usepackage{lipsum}

\title{Generating synthetic transactional profiles}

\author{
  Hadrien Lautraite\\
  Data Scientist\\
  National Bank of Canada\\
  Montreal \\
  \texttt{hadrien.lautraite@bnc.ca} \\
   \And
 Patrick Mesana \\
  Data Science Manager\\
  National Bank of Canada\\
  Montreal \\
  \texttt{patrick.mesana@bnc.ca} \\
}

\begin{document}
\maketitle

\begin{abstract}
Financial institutions use clients' payment transactions in numerous banking applications. Transactions are very personal and rich in behavioural patterns, often unique to individuals, which make them equivalent to personally identifiable information in some cases. \\
In this paper, we generate synthetic transactional profiles using machine learning techniques with the goal to preserve both data utility and privacy. A challenge we faced was to deal with sparse vectors due to the few spending categories a client uses compared to all the ones available. We measured data utility by calculating common insights used by the banking industry on both the original and the synthetic data-set. Our approach shows that neural network models can generate valuable synthetic data in such context.  Finally, we tried privacy-preserving techniques and observed its effect on models' performances.

\end{abstract}


\section{Introduction}

Organizations, such as financial institutions, want to create value with their data while preserving the privacy of individuals. The value of data can come from direct revenue streams or indirect ones by finding new insights for decision-making. It is often referred to the data utility. Banks use clients' transactions to empower their services and products through constant innovation, increasing transactional data utility but also increasing the risk of privacy leaks. De Monjoye et al. \cite{de2015unique} showed that in a transactional dataset of 1,1 million users of 3 months, only 4 transactions are needed to identify 90\% of clients. This is why there is a growing interest in privacy-preserving techniques for analytics and automation purposes.

Li, T. and Li, N showed that when you try to preserve data privacy, for example by applying generalization techniques on data, you reduce its utility \cite{li2009tradeoff}.
In the past, companies believed they could anonymize data while keeping an acceptable level of privacy and utility. Many private data-sets were anonymized and published on popular online platforms because of this belief. Researchers have shown many times now that anonymization is not without flaws and these now published data-sets can still contain personal information. A famous example is the re-identification attack on the Netflix Prize Dataset, published by Narayanan, A. and Shmatikov in 2008 \cite{narayanan2008robust}. 
Therefore, when managing data, a trade-off between utility and privacy needs to be recognized and addressed if one wishes to use privacy preserving techniques. 

A popular alternative to using data collected from people is to use synthetic data. A data is fully synthetic when it's not directly obtained from real data. In terms of privacy, it depends on how much the synthetic generative process is linked to real data. Thus, synthetic data do not eliminate privacy concerns and this has been studied since the introduction of fully synthetic data. \cite{rubin1993discussion} \cite{abowd2008protective}.  

One approach to generate synthetic data for analytics is to use simulators. Companies and researchers have been using them to test scenarios in their domain of expertise for a while. PaySim \cite{lopez2016applying} is a recent example of such simulator in the financial industry, it generates transactions focusing on specific use cases such as fraud detection. These simulators are often designed using statistics and subject matter experts. They have the advantage to be very loosely connected to real data and they have been proven to be useful. Their biggest drawback is that they are laborious to design. 
A less laborious approach is to use a generative model trained with machine learning techniques and this is the approach we chose in this work by using deep neural networks. The rise of big data makes them practical for many use cases. For example, Generative Adversarial Networks (GANs) \cite{goodfellow2014generative} have been used with great success to generate high dimensional data such as images.
Unfortunately, this approach is no silver bullet and a new type of inversion attack was found where attackers can recreate real faces only based on the model \cite{fredrikson2015model}.
Here again, there is a loss of privacy coming from the usage of real data to learn the weights of a Neural Network model.

In this paper, we recognize the utility-privacy trade-off by considering that any traceable connection to the real data is a potential leak of personal information. This can happen both in the learning phase and the validation phase. In the learning phase, the model has access to transactional data and possibly more information on individuals if auxiliary data are used (e.g. age, salary, address etc.). Because of this, it is possible that personal information or secrets are encoded in the weights of the Neural Network model \cite{carlini2019secret}. To diminish these memorization effects, we tried a Differential Privacy optimization technique that clips and adds noise to the gradient \cite{dwork2008differential} \cite{goodfellow2014generative}. In the validation phase, assuming the model's weights preserved privacy, the connection with real data depends on the architecture of the model. For two equivalent algorithms, one that depends on real data in validation preserves less privacy than one that do not use them as input.
For that reason, we went from using real data as input, to using auxiliary data instead, to not using data at all by generating synthetic data from noise. This method does not give any formal privacy guarantees but we believe it guided us towards more privacy-preserving models.

Our main focus was on measuring the utility of the models' output. Financial institutions have defined business use-cases that make the value analysis easier than evaluating the privacy risk. Therefore, in addition to the distance between the distributions of the real data and the synthetic data, we looked at the distance between spending categories histograms because we knew it is used by financial institutions. We also verified that sparsity was preserved because it is a key property of transactional data.

\section{Background: Generative models}

With the rise of deep learning, generative models have known great success. In particular variational auto-encoders and generative adversarial networks are two very popular and efficient ways to generate data. 

\subsection{Variational Auto Encoders}
Models in the auto-encoder family are composed of two neural networks trained simultaneously. The first model (encoder) maps input data to a latent space z, the decoder network maps the latent space to a reconstructed input. \par
For the specific case of variational auto-encoder \cite{Kingma2014AutoEncodingVB}, the latent space is  a distribution of the latent variables. The encoder learns parameters of a predefined distribution from which we sample values given to the decoder network.

\subsection{Generative Adverserial Network}
Generative Adversarial Networks (GAN) were introduced by Goodfellow and al \cite{goodfellow2014generative}. GAN are composed of two neural networks, a generator and a discriminator which are trained one against the other, hence the term adversarial. The generator is fed with a vector of random noise and generate an output vector with the same dimension as the real data. The Discriminator is trained, on real and generated data to distinguish between original data and data produced by the generator. 
Both models are trained one after the other, starting with the discriminator. During the training, the generator will learn to produce outputs more likely to fool the discriminator whereas the discriminator will become increasingly better at distinguishing between real and fake input. 

A common problem that can occur during training of a GAN is called mode collapse \cite{mode_collapse}. The generator specializes at generating one special kind of output which is good enough to foolish the discriminator. Eventually the discriminator will catch up and the generator will shift to producing outputs in another part of the output space. This leads to unstable training and poor results since most of the outputs are very similar. 
In this case, several points from the noise vector are mapped to the same region in the output space.

In this work we use two evolutions of the GAN classical architecture: Wasserstein GAN (WGAN)\cite{arjovsky2017wasserstein} and Conditional GAN (CGAN)\cite{mirza2014conditional}. 
Wasserstein GAN leverages the Wasserstein distance as a loss for the GAN. WGAN are less subject to the mode collapse issue and are known for having better convergence properties. In addition, instead of measuring the ability of the generator to fool the discriminator, the Wasserstein loss gives us a direct indication of the quality of the synthetic data. The discriminator becomes a critic that gives a score measuring realness or fakeness of a given input vector.

Conditional Generative adversarial neural networks (CGAN) are another evolution of GAN. In addition to the random noise, the generator is fed with some auxiliary information. This auxiliary information is also passed to the discriminator/critic model. The idea behind this is generating synthetic data coherent with the auxiliary data. Hence it can be used to avoid mode collapse. Noise combined with different auxiliary data should be mapped to a different part of the output space. 

Finally, we combine the ideas behind Conditional and Wasetstein GAN. Using a conditional GAN with a Wasserstein loss. 

\begin{figure}[h]
    \includegraphics[width=10.cm]{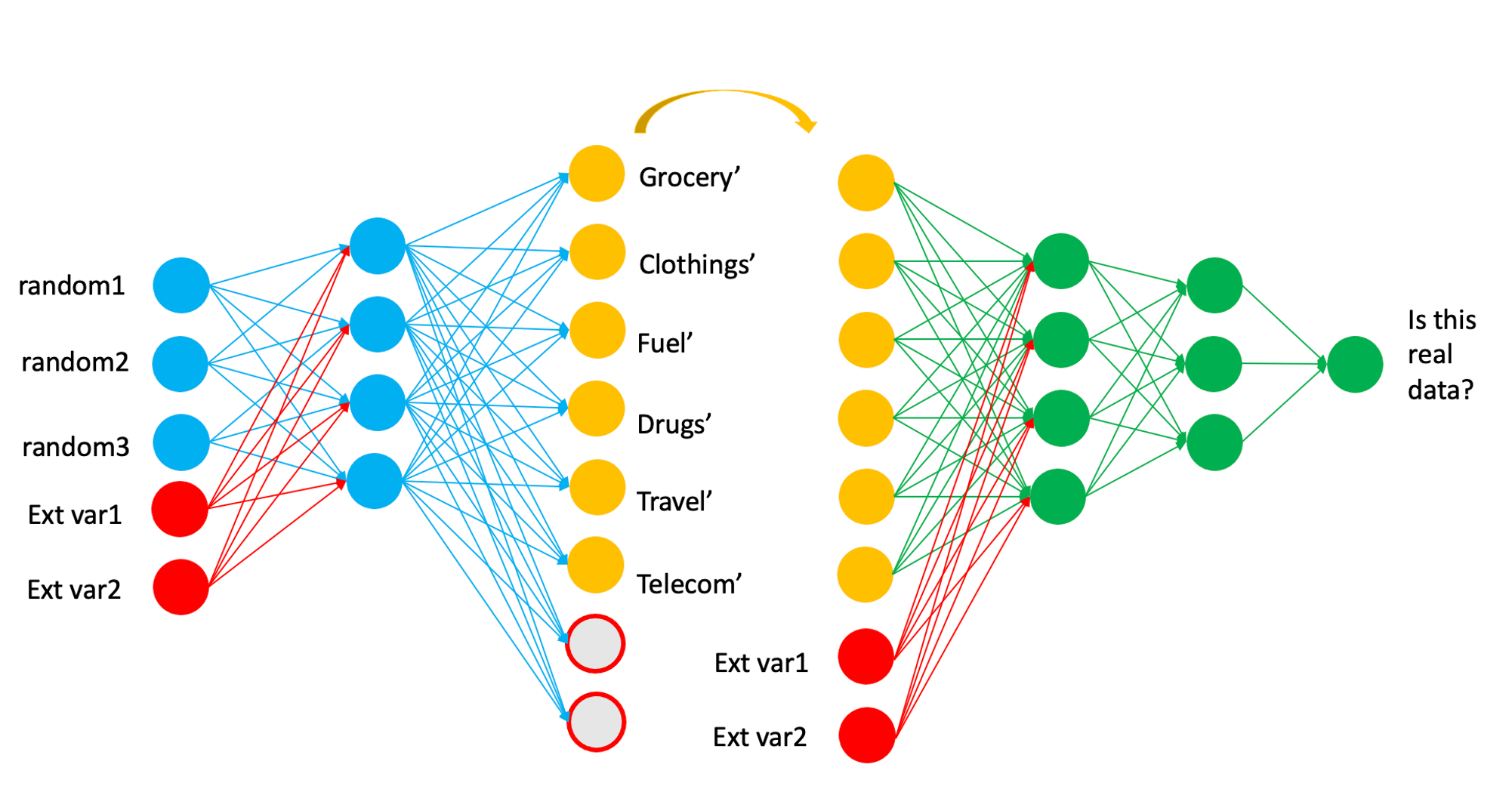}
    \centering
    \caption{Architecture of our Conditional GAN models}
    \label{fig:CGAN}
\end{figure}

\section{Dataset}

Credit card transactional data are very sensitive. Even if the unique identifier (account number or client number) and other personally identifiable information (PII) are removed, transactions made by someone create a footprint that make re-identification quite easy. According to Yves-Alexandre de Montjoye et al \cite{de2015unique}, 90\% of the population can be uniquely identified with as few as 4 transactional data points. Moreover, those data are considered very personal by most people. Meanwhile, transactional data are widely used in banks and have numerous applications, from identifying active customers to fraud detection. 
The combination of the high value of those data and their extreme sensitivity makes them a prime candidate for privacy preserving data analysis and synthetic data generation. 

For our work we have been using a dataset of credit card transactional profiles. 
We define transactional profiles as vectorized data transactions. Each vector represents the amount of money spent among different purchasing categories (SIC code) for a given month. 

People often have multiple credits cards, thus the information detained by the bank is incomplete and many transactional profiles are empty or nearly empty. We decided to focus on clients that are actively using their credit cards, making the transactional profiles more meaningful. We filtered our dataset by only keeping clients that spend more than a certain amount in total in the sum of the top 26 most common categories. 
The resulting dataset has around 300 000 observations which are split randomly (80\%-20\%) between train and testing set. 

Another particularity of the dataset is its sparseness. Most of the column have a value of 0\$ spent in the category for a given month. We cannot ignore those columns as the presence of a value different from 0 bears a lot of information, even if infrequent. As an example, if we consider the category: spending in airplane tickets, most of the clients will not spend any money in this category during a typical month. But for the clients that do, it brings useful information that can be used by financial institutions to offer custom services to customer. Finally, the dataset contains a significant number of outliers, clients that will spend a very high amount of money for a given category in comparison to the rest of the population. For those reasons we used a log transformation of the dataset, taking the log of the amount spent per category.

\section{Methodolgy}

Our goal is to generate synthetic transaction profiles that could be used for real business use cases. We compare different approaches to generate synthetic data with different levels of dependencies to the real data for data generation.
We start with models that use real transactional data for both training and generation of new transactional profiles: VAE trained on real data and fed with data for generation. 
We then use models that rely on real transactional and auxiliary data for training and only on real auxiliary data and a vector of random noise for generation of new observations: C-GAN and WC-GAN. 
Finally, models that only depend on real transactional data during the training phase: VAE trained on real data and data generation from random noise.


In order to compare the approaches and the models, we focus on 3 criteria. First, we will visually inspect the synthetic data distribution to see whether or not the generated data are sparse. We look at the sparseness in two different manners, first most of the values for a specific individual should be zero: during a month a client will spend money with his credit card in some categories but not all. Second, we look at the sparsity for every feature: typically for a given category we expect that only a given number of clients will make transactions, with the rest spending zero dollars for goods or services belonging to this category. Even if this criterion might seem naive, it actually has value from a business perspective: knowing the percentage of customers using their credit card for some specific categories (e.g.: grocery) gives the company an idea of the percentage of clients using our product for their primary needs. \par
Then we look at the Wasserstein distance between synthetic and real data. We calculate the distance between the two dataset for each input variable and average of all distances as a metric. With this criterion we measure how good the model is at generating data for a given column.
Knowing the distribution of the amount spent for a specific category is a valuable information for the bank and can be considered as an insight. By measuring the distance between the two distributions, we obtain a measure of utility. How close the insights taken from the real data are to the synthetic ones.
Finally, we compare the cosine distance (1-cos) between pairwise observations. It measures how close the transactional profile of one specific customer is to a synthetic one. For business needs, not only we want the synthetic data to be close to the original data for one specific feature across the whole population but we also want transactional profiles to be coherent across different features for one individual. The cosine distance allows us to compare the amount of money spent in all categories for a real and a synthetic customer. 
In order to measure the cosine distance, we need to generate data that are similar to paired reference 
real customers.

For the VAE we use two different data generation techniques, that have different levels of dependencies with regards to the real data. 
First, we feed real data to the Encoder and sample a new representation in the latent space. 
We then pass the obtained vector to the decoder of the VAE to get our reconstructed transactional profiles. Data obtained with this method are close to the original input. Even if there is a random sampling mechanism taking place in the VAE's latent space, one could argue they are not new synthetic observations but rather reconstruction of the original input.
For the second approach we generate normally distributed (mean:0, standard deviation:1) random vectors with the same shape as the latent space. We pass those vectors to the decoder  already trained on real data to generate synthetic data.

For the C-GAN and WC-GAN we use socio-demographics information about the client (age, salary, credit limit) and statistics about purchases (credit card balance, last repayment amount) as conditional auxiliary information. Models are trained with real transactional data and real auxiliary data. Then we generate new observations using real auxiliary information in addition to a vector of random noise.

All our experiments were performed on a spark cluster, with most of our model training taking place on the driver node. We did not use GPU for our experiments. As a consequence, we did not spend much time doing complex hyperparameters search. We only tried a few different architectures.

\section{Results}

In this section, we analyze the results obtained with the different generation settings:
real transactional data for training and generation (we name this model VAE from data), real transactional data and real auxiliary data for training and real auxiliary data plus random noise vector for generation (C-GAN, WC-GAN) and finally, real transactional data for training, and data generated from noise only (VAE from noise)

\subsection{With all the spending categories}

First, we analyze the results on the whole dataset.\par

As expected, the VAE from data model gets good results. This is expected since it is the model that depends the most on data. The model gets the best performances in term of Wasserstein distance.\par
Both GAN models (C-GAN and WC-GAN) succeed at generating sparse data, keeping this key characteristic of the dataset. The Conditional GAN model gets the best performances between the two models both in term of Wasserstein distance and cosine distance (best overall). \par
We notice that WC-GAN models tend to produce some extreme values with spending over one billion dollars for a single SIC code. Even if the frequency of these extreme values is low it has a huge impact on the Wasserstein distance. It is interesting to notice that these rare high value spending behaviors exist in the original dataset but with a lower magnitude (some tens of thousands of dollars).\par
Even though Wasserstein distance seems to be reasonably small between true data and data generated from the VAE from rand model, the model fails to produce sparse data. In term of utility, the sparseness is a key component. We can consider that the data produced by this model brings less analytical value.\par

Table \ref{tabl:perf_full} shows the metrics evaluated on the entire dataset.

\begin{table}[h]
\begin{tabular}{|l|l|l|l|}
\hline
\textbf{Model}   & \textbf{Sparsity of generated data} & \textbf{Average Waserstein distance} & \textbf{average cosine distance}  \\ \hline
VAE - from data  & yes             & 0.5                                  & 0.13                                                        \\ \hline
VAE - from noise & no              & 2.4                                  & 0.24                                                        \\ \hline
CGAN             & yes             & 2.66                                 & 0.07                                                          \\ \hline
WCGAN            & yes             & 423.41                               & 0.19                                                          \\ \hline
\end{tabular}
\caption{Model performances based on our three evaluation criteria}
\label{tabl:perf_full}
\end{table}

The VAE from data model creates synthetic data with a distribution of the amount of money spent for a given spending category similar to the real data. The C-GAN model is able to generate well individual observations that are similar to the real individuals.

\subsection{Keeping only the most important spending categories}
We also look closer at the results on a set of categories which are the most used.
The VAE from data model gets the best performance in term of Wasserstein distance and the C-GAN gets the best cosine distance. 

Table \ref{tabl:perf_mc} summarizes the results on the most commonly used categories.

Given real auxiliary information, the conditional GAN is able to generate a synthetic transactional data vector with an average 26  degrees angle between synthetic and original data for most common spending categories.

\begin{table}[h]
\begin{tabular}{|l|l|l|l|}
\hline
\textbf{Model}   & \textbf{Sparsity of generated data} & \textbf{Average Waserstein distance} & \textbf{average cosine distance} \\ \hline
VAE - from data  & yes             & 2.27                                 & 0.19                                                         \\ \hline
VAE - from noise & no              & 23.05                                & 0.30                                                         \\ \hline
CGAN             & yes             & 24.84                                & 0.10                                                        \\ \hline
WCGAN            & yes             & 30.26                                & 0.24                                                         \\ \hline
\end{tabular}
\caption{Model performances based on our three evaluation criteria on the most common variables}
\label{tabl:perf_mc}
\end{table}

Finally, we compare the percentage difference between an insight computed on real data and one computed on synthetic ones. Our insights are: the percentage of people that spend at least 1\$ in groceries, the percentage of people that spend at least 10\$ in clothing and the percentage of people that spend at least 100\$ in the sum of the 20 most common spending categories. 
Insights computed on the data generated by the VAE from data model are the closest to the reality. Conditional GAN produces insights quite close to the real ones for small spending. Indeed, the C-GAN model mainly produces small values. Its performances decrease quickly when when the spending threshold in dollars increases. The WC-GAN model suffers less from this issue and produces insights relatively close to the ones computed on the real dataset even with a larger spending threshold. 
Table \ref{tabl:spent_sic} shows the difference in insights computed using real and generated dataset.

\begin{table}[h]
\begin{tabular}{|l|l|l|l|}
\hline
\textbf{Model}   & \textbf{grocery spendings \textgreater{}1\$} & \textbf{clothing spendings \textgreater 10\$} & \textbf{most common spendings \textgreater 100\$} \\ \hline
VAE - from data  & 0                                            & 0                                             & 0                                                 \\ \hline
CGAN             & 0.05                                         & 0.22                                          & 0.91                                              \\ \hline
WCGAN            & 0.26                                         & 0.06                                          & 0.35                                              \\ \hline
\end{tabular}
\caption{Absolute value of the difference in the proportion of people that spent at least X\$ for a specific SIC code between real and synthetic data}
\label{tabl:spent_sic}
\end{table}

To sum up, the model which depend the most on data: VAE from data produces the most convincing input variable distributions. This model gets the best performances in term of Wasserstein distance and closeness of insights. 
Conditional GAN produces  interesting results at the individual observation level, with an average angle of 21.5 degrees between a synthetic and a real transactional vector with all the spending categories given the real auxiliary information and a vector of random noise. However, even if the model produces sparse data, the predicted amount of money spent per SIC has often a value lower than reality. The WC-GAN model suffers from an opposite problem and produces rare extreme values that have a negative impact on the Wasserstein distance. 

\subsection{Note on Differential Privacy}
In our experiments we fail to train GAN using differential private gradient decent. We adopt an approach similar to \cite{torkzadehmahani2019dp} applying noise and clipping the gradient during the discriminator training.  The data generated with DP are not a good representation of the real data and the models fail to produce sparse data.
It should be noted that the VAE models are the ones that suffer the least from training with differential private SGD.

\section{Conclusion and Further Work}

In this research, the strongest candidate to generate profiles according to our utility metrics was the Variational Auto Encoder (VAE) using real data as input for data generation. This dependency on real data after training the model, makes us question if we can consider the generated data as synthetic. We showed that Conditional Generative Adversarial Network, compared to other models, is also a performant model to generate sparse transactional profiles. It is not using real transactional profiles after training and it uses auxiliary data instead. We believe this approach offers privacy benefits, although we know there is a siginificative correlation between transactions and socio-demographics. This correlation can be used by identity thiefs to design re-identification attacks, they narrow people in a database using auxiliary information they can find on them. In a further work, we should study these potential attacks on synthetic data and we should find privacy metrics to compare models that use auxiliary data for data generation with models that do not use them.

The experiments using Stochastic Gradient Descent with Differential Privacy to reduce memorization effects were non conclusive. We found it was not easy to parameter the library we used (Tensorflow-privacy). Consuming only a small amount of privacy budget per iteration affected our models' performance by a lot and we could not identify the reason. We should certainly investigate more in the future.

Finally, we found that measuring utility by focusing on insights financial institutions use, refined our analysis compared to only using point by point distance metrics such as cosine. Improvement in one does not always imply improvement on the other. Using noise to generate data from the VAE showed that the distance between spending categories histograms was acceptable but sparsity was lost. 
To continue our utility analysis, we think we should define more business metrics based on the data-set. We also want to add a machine learning task used in a real business case to expose more complex usage of the synthetic data.

\section{Acknowledgement} 
We would like to warmly thank: Nada Naji, Reyhaneh Rezvani and Eric Charton for their re-reading of this article and their precious advice.

\bibliographystyle{unsrt}  
\bibliography{article}






\end{document}